\documentclass{article}

\usepackage{amssymb,amsthm,amsmath,array}
\usepackage{appendix}
\usepackage{graphicx}
\usepackage[footnotesize]{caption}
\usepackage[caption=false,font=footnotesize]{subfig}

\usepackage{xspace}
\graphicspath{{Figures/}}
\usepackage[sort&compress, numbers]{natbib}
\usepackage{stmaryrd}
\usepackage{xcolor}

\usepackage{tikz}
\usetikzlibrary{arrows}

\usepackage{pgfplots}
\usepackage{multirow}
\definecolor{mycolor1}{rgb}{0.83,0.83,1}

\newtheorem{thm}{Theorem}

\newtheorem{prop}{Proposition}

\newtheorem{remark}{Remark}
\usepackage{color}
\usepackage[hidelinks]{hyperref}

\newcommand{\jth}{{^{\rm th}}}

\newcommand{\Nxi}{{N_\xi}}
\newcommand{\Np}{{N_{\rm p}}}
\newcommand{\Nhs}{{N_{\rm hs}}}
\newcommand{\Nyq}{{\rm nyq}}
\newcommand{\range}[1]{\llbracket #1 \rrbracket}

\newcommand{\supp}{{\rm supp}\,}

\newcommand{\st}{{\rm s.t.}}

\DeclareMathOperator*{\argmin}{arg\,min}

\newcommand{\bs}{\boldsymbol}
\newcommand{\bb}{\mathbb}
\newcommand{\cl}{\mathcal}

\newcommand{\ts}{\textstyle}
\newcommand{\ie}{\emph{i.e.},\xspace}
\newcommand{\eg}{\emph{e.g.},\xspace}

\newcommand{\nell}{{l}}

\newcommand{\iid}{%
  \ifmmode
  \mathrm{i.i.d.}%
  \else%
  i.i.d.\@\xspace%
  \fi%
}
\newcommand{\rv}{\mbox{r.v.}\xspace}

\newcommand{\sq}{\vspace{-2mm}}

\usepackage[top=1in, left=1in, right=1in, bottom=1in]{geometry}
\renewenvironment{abstract}{\bf {\em\ Abstract---}}{}

\makeatletter
\renewcommand{\section}{\@startsection {section}{1}{\z@}%
             {-3.5ex \@plus -1ex \@minus -.2ex}%
             {2.3ex \@plus.2ex}%
             {\normalfont\large\bfseries}}
\makeatother
\begin{document}

\title{Compressive Single-pixel Fourier Transform Imaging\\ using Structured Illumination}
\author{A. Moshtaghpour$^1$, J. M. Bioucas-Dias$^2$, and L. Jacques$^1$\footnote{AM is funded by the FRIA/FNRS. LJ is funded by the F.R.S.-FNRS.}\\
  \footnotesize $^1$ ISPGroup, ICTEAM/ELEN, UCLouvain, Louvain-la-Neuve, Belgium.\\[-1mm]
  \footnotesize $^2$ Instituto de Telecomunica\c c\~oes, Instituto Superior T\'ecnico, Universidade de Lisboa, Portugal.} \date{\empty}

\maketitle

\begin{abstract}
Single Pixel (SP) imaging is now a reality in many applications, \textit{e.g.,} biomedical ultrathin endoscope and fluorescent spectroscopy. In this context, many schemes exist to improve the light throughput of these device, \textit{e.g.,} using structured illumination driven by compressive sensing theory. In this work, we consider the combination of SP imaging with Fourier Transform Interferometry (SP-FTI) to reach high-resolution HyperSpectral (HS) imaging, as desirable, \eg in fluorescent spectroscopy. While this association is not new, we here focus on optimizing the spatial illumination, structured as Hadamard patterns, during the optical path progression. We follow a variable density sampling strategy for space-time coding of the light illumination, and show theoretically and numerically that this scheme allows us to reduce the number of measurements and light-exposure of the observed object compared to conventional compressive SP-FTI.
\end{abstract}


\section{Introduction}                           
\label{sec:intro}           
Nowadays, Single Pixel (SP) imaging has become an emerging paradigm for capturing high quality images using a single photo detector \cite{duarte2008single,sivankutty2018nonlinear,guerit2018,davis2004hyperspectral}, in a low cost and high light throughput process with low memory requirement. These advantages have made SP imaging an excellent candidate for recording HyperSpectral (HS) volumes~\cite{roman2014asymptotic,studer2012compressive}, in particular, when it is combined with Fourier Transform Interferometry (FTI)~\cite{jin2017hyperspectral,moshtaghpour2018compressive}.

\begin{figure}
\centering
\includegraphics[width = 1\columnwidth]{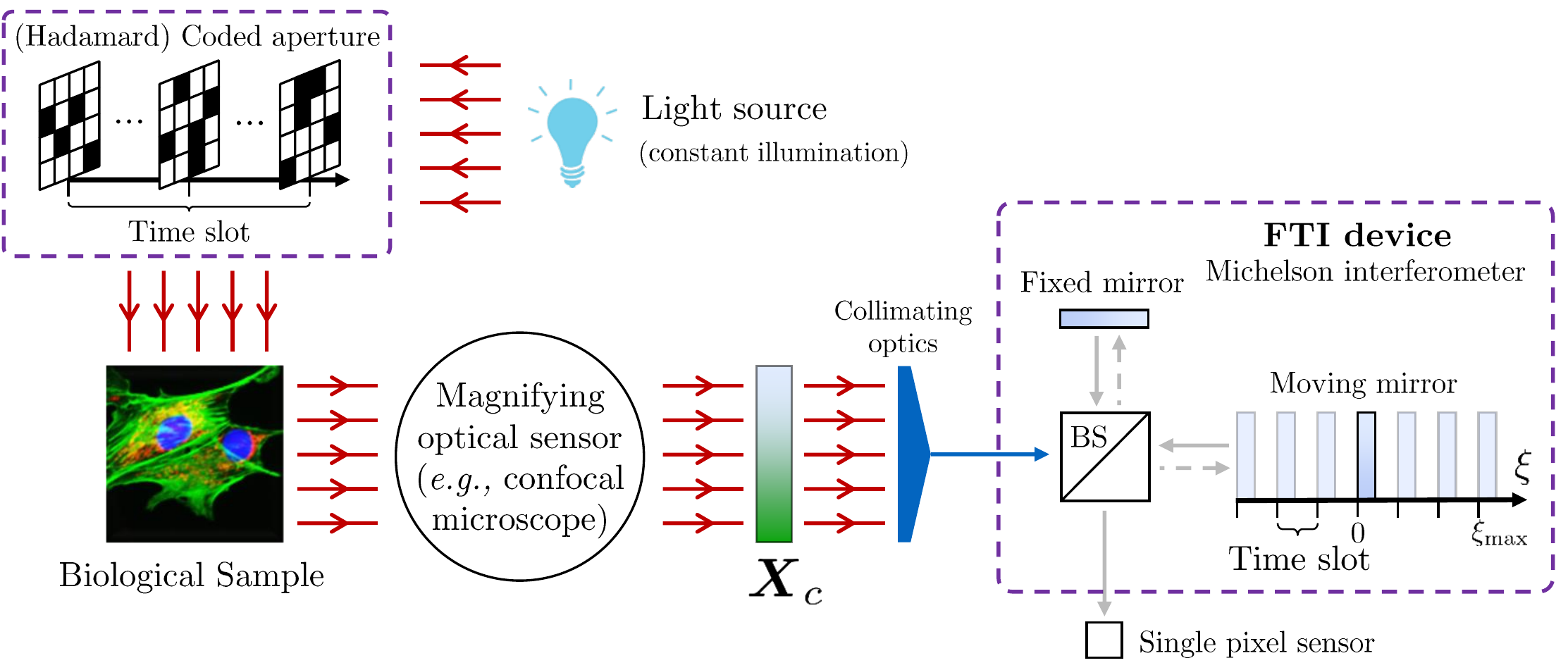}
\caption{Operating principles of SP-FTI. A continuous HS volume $\bs X_c$ (\eg observed from a confocal microscope) is spatially and temporally modulated from space-time coding of the light source.}
\sq \sq \sq
\label{fig:FTI_scheme}
\end{figure}

As schemed in Fig.~\ref{fig:FTI_scheme} (dashed box on the right), FTI works on the principle of a Michelson interferometer. A coherent wide-band beam entering the FTI device is first divided into two beams by a Beam-Splitter (BS). Those beams are then reflected back either by a fixed mirror or by a moving mirror, controlling the Optical Path Difference (OPD) of the two beams, and interfere after being recombined by the BS. The resulting beam is later recorded by an external imaging sensor. Physical optics shows that the outgoing beam from the FTI, as a function of OPD $\xi\in \bb R$, is the Fourier transform of the entering beam, as a function of wavenumber $\nu \in \bb R$. As an advantage, the spectral resolution of the HS volume can be increased by enlarging the range of recorded OPD values. However, in biomedical applications, this increase of resolution is limited by the tolerance of the biological elements against the light exposure.

On the other hand, the theory of Compressed Sensing (CS) \cite{donoho2006compressed,candes2006near,foucart2013mathematical} has shown successful results \cite{moshtaghpour2016,moshtaghpour2017a,moshtaghpour2018multilevel,moshtaghpourcoded} in reducing the amount of light exposure during the FTI acquisition, while the spectral resolution is preserved. We have proposed two compressive sensing-FTI methods in \cite{moshtaghpour2018}, \textit{i.e.,} Coded Illumination-FTI (CI-FTI) and Structured Illumination-FTI (SI-FTI), that can efficiently reduce the light exposure on the observed object. CI-FTI operates by temporal coding of the global light source, while in SI-FTI, spatial modulation of the illumination (\eg by a spatial light modulator), allows for different OPD coding per spatial locations. 

Unlike the above-mentioned works, we here propose an FTI-based HS acquisition, that is constrained to use SP imaging. In the corresponding device, referred to as Single Pixel Fourier Transform Interferometry (SP-FTI), the light distribution is coded (or structured) in both OPD (or time) and spatial domains before being integrated into a single beam (unlike SI-FTI), using, \eg collimating optics (see Sec.~\ref{sec:SP-FTI}). While SP-FTI has been implemented in food monitoring application \cite{jin2017hyperspectral}, its theoretical analysis is not covered in the literature.

In this paper, adopting the stable and robust sampling strategies of Krahmer and Ward \cite{krahmer2014stable} for compressive imaging, we optimize SP-FTI sensing by following a Variable Density Sampling (VDS) of both the OPD domain and the Hadamard transformation of the spatial domain. This differs from a former work~\cite{moshtaghpour2018compressive} where we considered multilevel sampling. In particular, our sampling strategy relies on the estimation of tight bounds on the local coherence between the 3D sensing and sparsity bases. The sensing basis is obtained from the Kronecker product of the Fourier (imposed by the FTI system) and Paley Hadamard (an aspect of our sensing design) bases and the sparsity basis is achieved by the Kronecker product of the 1D Haar wavelet and the 2D isotropic Haar wavelet bases. Our analysis in Appendix is versatile and could be extended to other SP imaging techniques.

The rest of the paper is structured as follows. We first summarize the recovery guarantee associated with VDS theory in Sec.~\ref{sec:compr-sens}. After describing the principles of SP-FTI, our stable and robust compressive SP-FTI is proposed in Sec.~\ref{sec:main-results}. Numerical simulations are provided in Sec.~\ref{sec:numerical-results}.

\textbf{Notations}:
domain dimensions are represented by capital letters, \eg $K,M,N$. Vectors, matrices, and data cubes are denoted by bold symbols. 
For any matrix (or vector) $\bs V \in \bb C^{M\times N}$, $\bs V^\top$ and $\bs V^*$ represent the transposed and the conjugate transpose of $\bs V$, respectively, and $\bs{V} \otimes \bs{W}$ denotes the Kronecker product of two matrices $\bs V$ and $\bs W$. The $\ell_p$-norm of $\bs{u}$ reads $\|\bs{u}\|_p := (\sum_i |u_i|^p)^{1/p}$, for $p\geq 1$, with $\|\bs u\| :=\|\bs u\|_2$. The identity matrix of dimension $N$ is represented as $\bs{I}_N$. Finally, we use the asymptotic relations $f \lesssim g$ (or $f \gtrsim g$) if $f \le c\,g$ (resp. $g \le c\, f$) for two functions $f$ and $g$ and some value $c>0$ independent of their parameters. 
\section{Compressed Sensing with random partial Orthonormal Bases}          
\label{sec:compr-sens}
One interest of CS theory is to recover a signal $\bs x \in \bb C^{N}$ from a vector of noisy measurements~\cite{candes2007sparsity}
\begin{equation}\label{eq:cs-model}
\small \bs y = \bs P_\Omega\bs \Phi^* \bs x + \bs n,
\end{equation}
where $\Omega = \{\omega_j\}_{j=1}^{M} \subset \range{N}$ is a multiset of (non-unique) indices selected from a sampling strategy $p$, \ie these indices are independently and identically distributed (\iid) according to the probability mass function (pmf) $p$ on $\range{N}$ such that $p(l) = \bb P[\omega_j=l]$ for any $l \in \range{N}$ and $j \in \range{M}$.  The restriction operator is denoted by $\bs P_\Omega \in \bb \{0,1\}^{M \times N}$ with $(\bs P_\Omega \bs x)_j = x_{\omega_j}$ and $\bs n$ models an additive noise. If $\bs x$ is assumed sparse (or well approximated by a sparse representation) in some basis $\bs \Psi$, \ie $\bs x = \bs \Psi \bs s$, with $K:=|\supp (\bs s)| \ll N$, then the following proposition, which is a direct combination of \cite[Thm.~5.2]{krahmer2014stable} and \cite[Prop.~3]{moshtaghpour2018}, provides the conditions and guarantee under which $\bs x$ can be estimated from the noisy measurements $\bs y$.
\begin{prop}[VDS guarantee]\label{prop:vds}
Let $\bs \Phi, \bs \Psi \in \bb C^{N \times N}$ be orthonormal sensing and sparsity bases, respectively. We assume a local coherence $\mu_l$ bounded as
\begin{equation}\label{eq:local-coherence}
\mu_l(\bs \Phi^* \bs \Psi) := \ts \max_{j} |(\bs \Phi^* \bs \Psi)_{l,j}| \le \kappa_l,~~l\in\range{N}.
\end{equation}
for some $\kappa_l >0$, and $\bs \kappa\!:= \![\kappa_1,\cdots,\kappa_N]^\top$. Suppose $\epsilon\! \in\! (0,1]$, $K \gtrsim \log(N)$, 
\begin{equation}\label{eq:sample-complexity}
M \gtrsim \|\bs \kappa\|^2 K \log(\epsilon^{-1}),
\end{equation}
and generate $\Omega =\{\omega_1,\cdots,\omega_M\}\subset \range{N}$ as explained above according to the pmf
\begin{equation}\label{eq:pmf-vds}
p(l):=\kappa_l^2/\|\bs \kappa\|^2.
\end{equation}
Consider the diagonal matrix $\bs D \in \bb R^{M \times M}$ with $d_{jj} = 1/\sqrt{p(\omega_j)}$, $j \in\range{M}$. If $\bs x \in \bb C^{N}$ is a signal observed by the noisy sensing model \eqref{eq:cs-model} with $\|\bs D \bs n\|\le \varepsilon$, then, the solution $\hat{\bs x}$ of the program 
\begin{equation}\label{eq:bpdn}
\Delta(\bs y,p,\Omega,\bs \Phi,\bs \Psi)\! := \\ \! \argmin_{\bs u \in \bb C^{N}}\! \|\bs \Psi^*\bs u\|_1 ~\st ~\|\bs D(\bs y-\bs P_{\Omega}\bs \Phi^* \bs u)\|\!\le\! \varepsilon,
\end{equation}
 satisfies $\|\bs x - \hat{\bs x}\| \le 2 K^{-1/2} \sigma_K(\bs \Psi^* \bs x) + \varepsilon$, with probability exceeding $1-\epsilon$, where $\sigma_K(\bs u):=\|\bs u -\cl H_K(\bs u)\|_1$ is the best $K$-term approximation error and $\cl H_K$ is the hard thresholding operator that maps all but the $K$ largest magnitude entries of the argument to zero.
\end{prop}
\section{From Nyquist to Compressive SP-FTI}             
\label{sec:SP-FTI}
Let us first study a Nyquist implementation of SP-FTI, \ie collecting as many observations as the number of HS volume voxels, and where HS reconstruction can be achieved by a linear (inverse) transform. Let $\bs X \in \bb R^{N_\nu \times \Np}$ be the discretization of the HS volume $\bs X_c$ (see Fig.~\ref{fig:FTI_scheme}) over $N_\nu = N_\xi$ wavenumber samples and $\bar{N}_{\rm p}$ pixels in each $\rm x$- and $\rm y$-axis such that $N_{\rm p}:=\bar{N}_{\rm p}^2$. We assume that the light source provides constant illumination during $\Nhs := N_\xi\Np$ time slots (associated to $\Nxi$ OPD samples). As shown in Fig.~\ref{fig:FTI_scheme} (top-left), at each OPD sample $l_\xi \in \range{N_\xi}$, we consider a Coded Aperture (CA) that is programmed $\Np$ times with all columns of the Hadamard basis $\bs \Phi_{\rm had} \in \{\pm 1/\sqrt{\Np}\}^{\Np \times \Np}$ (up to a rescaling, see Remark~\ref{remark:on-actual-spfti-model})~\cite{horadam2012hadamard}. Every programmed CA gives a spatially coded HS light beam, which is later integrated into a single beam, \eg by means of an optical collimator. Following the principles of the FTI presented in Sec.~\ref{sec:intro}, the light sensitive element records one Fourier coefficient (corresponding to the current OPD sample) of the spectral energy of the entering beam.

Mathematically, the $l\jth$ time slot corresponds to the $l_{\rm p}\jth$ Hadamard pattern programmed at $l_\xi\jth $ OPD point such that $l = \Np (l_\xi -1) + l_{\rm p} \in \range{\Nhs}$, $l_\xi \in \range{\Nxi}$, and $l_{\rm p} \in\range{\Np}$. Therefore, each SP-FTI observation reads
\begin{equation*}\label{eq:spfti single acquisition}
\small y^\Nyq_l  = \bs P_{\{l_\xi\}} \bs \Phi^*_{\rm dft} \bs X \bs \Phi_{\rm had} \bs P^\top _{\{l_{\rm p}\}} + n^\Nyq_l= \bs P_{\{l\}} \bs \Phi^*_{\rm sp} \bs x + n^\Nyq_l,
\end{equation*}
where $\bs x := \rm vec(\bs X)$, $\bs \Phi_{\rm dft} \in \bb C^{\Nxi \times \Nxi}$ denotes the 1D discrete Discrete Fourier Transform (DFT) basis, $\bs \Phi_{\rm sp}:= \bs \Phi_{\rm had} \otimes \bs \Phi_{\rm dft}$ and $n^\Nyq_l$ models a noise, which is assumed here to be an additive i.i.d. Gaussian noise, \ie $n^\Nyq_l \sim_\iid \cl N(0,\sigma^2_{\Nyq})$. By collecting all the observations in the vector $\bs y^\Nyq :=[y_1^\Nyq,\cdots,y_{\Nhs}^\Nyq]^\top$, the whole SP-FTI mixing model can be written as
\begin{equation}\label{eq:spfti full acquisition}
\small \bs y^\Nyq=  \bs \Phi^*_{\rm sp} \bs x +\bs n^\Nyq.
\end{equation}
\begin{remark}\label{remark:on-actual-spfti-model}
As the Hadamard matrix contains $\pm \Np^{\rm -1/2}$ entries, it has to be properly transformed via $\bs \Phi_{\rm bin} := (\sqrt{\Np}\bs \Phi_{\rm had}+\bs 1_{\Np} \bs 1^\top_{\Np})/2 \in \{0,1\}^{\Np \times \Np}$ so that it can be implemented with 0's and 1's on a CA. However, the measurements \eqref{eq:spfti full acquisition} can be extracted from the actual measurements acquired by $\bs \Phi_{\rm bin}$, as long  as the all-on CA (associated with the first row of Hadamard matrix) is considered in the acquisition process \cite{sudhakar2015compressive,roman2014asymptotic,studer2012compressive}.
\end{remark}
We now propose a compressive form of SP-FTI, where only $M$ Hadamard patterns (out of $\Nhs$ possible patterns) are programmed, \ie we assume that CA blocks out the light illumination during the other $\Nhs - M$ time slots. The compressive SP-FTI system can be modeled as
\begin{equation}\label{eq:spfti cs acquisition}
\small \bs y^{\rm sp}=  \bs P_\Omega \bs \Phi^*_{\rm sp} \bs x +\bs n \in \bb R^M,
\end{equation}
where $\bs n = [n_1,\cdots,n_M]^\top$ with $n_l \sim_\iid \cl N(0,\sigma^2_{\Nyq})$ and $\Omega = \{\omega_j\}_{j=1}^{M} \subset \range{\Nhs}$ is the subsampled index set of Hadamard patterns. Considering Remark~\ref{remark:on-actual-spfti-model}, a careful analysis reveals that the total light exposure received by the observed object during compressive SP-FTI acquisition is proportional to $(M+\Nxi)\Np$, compared to $(\Nhs + \Nxi)\Np$ in Nyquist SP-FTI, where the term $\Nxi$ is induced by Remark~\ref{remark:on-actual-spfti-model}. The light exposure can thus be reduced by a factor of $(M+\Nxi)/(\Nhs+\Nxi)$. In the next section, we optimize the pmf of each $\omega_j$ according to a sparsity model of $\bs x$, so that the number of required measurements $M$ is minimized. This differs from~\cite{jin2017hyperspectral} where a fixed number of Hadamard patterns, chosen uniformly at random, are programmed at every OPD point.
\section{Main Results: Stable and Robust Compressive SP-FTI}             
\label{sec:main-results}
W now provide a stable and robust scheme to recover every HS volume from our compressive SP-FTI system by following the general VDS scheme of Sec. \ref{sec:compr-sens}, with special care to integrate the spatiospectral geometry of SP-FTI.

As any other CS applications, HS data recovery from under-determined measurements in \eqref{eq:spfti cs acquisition} requires an accurate low-complexity prior model on the HS volume. Our observations confirm that biological HS data commonly observed in FS share sparse/compressible representation in the Kronecker product of the 1-D Discrete Haar Wavelet (DHW) basis $\bs \Psi_{\rm dhw}$ and the 2-D Isotropic Discrete Haar Wavelet (IDHW) basis $\bs \Psi_{\rm idhw}$. According to Prop.~\ref{prop:vds}, given the noisy compressive SP-FTI measurements in \eqref{eq:spfti cs acquisition}, a stable and robust HS volume recovery can be achieved by solving $\Delta (\bs y^{\rm sp},p,\Omega,\bs \Phi_{\rm sp},\bs \Psi_{\rm sp})$ in \eqref{eq:bpdn}, where $\bs \Psi_{\rm sp} := \bs \Psi_{\rm idhw} \otimes\bs \Psi_{\rm dhw}$ denotes a 3-D analysis sparsity basis.

We can now adjust the optimum pmf determining $\Omega$ from~\eqref{eq:pmf-vds}. Given $ \kappa_l \ge \mu_l(\bs \Phi^*_{\rm sp}\bs \Psi_{\rm sp})$, for $l \in \range{\Nhs}$, Prop.~\ref{prop:vds} shows that, for a fixed $K$ and $\epsilon$, we must record $M \gtrsim \|\bs \kappa\|^2 K \log(\epsilon^{-1})$ SP-FTI measurements w.r.t. the pmf $p(l) = \kappa^2_l/\|\bs \kappa\|^2$ with $l \in \range{\Nhs}$.  We prove in Appendix that, in the context of SP-FTI,
\begin{equation}\label{eq:local-coherence-spfti}
\small \kappa_l := \sqrt{2} \min \{1,2^{-\lfloor \log_2(\max\{l_{\rm x},l_{\rm y}\}-1)\rfloor} \cdot |l_\xi-\Nxi /2|^{-1/2}\},
\end{equation}
 and $\|\bs \kappa\|^2 \lesssim \log(\Nxi) \log(\Np)$, where $l = l(l_\xi,l_{\rm x},l_{\rm y}) := \Np (l_\xi -1)+ \bar{N}_{\rm p}(l_{\rm y}-1)+l_{\rm x}$, relates a 1D index $ l\in \range{\Nhs}$ to a 3D index representation $(l_\xi,l_{\rm x},l_{\rm y})$ where $l_\xi \in \range{\Nxi}$ denotes the OPD index and $l_{\rm x}, l_{\rm y} \in \range{\bar{N}_{\rm p}}$ are the spatial Hadamard ``frequencies''. In this case, \eqref{eq:pmf-vds} in Prop.~\ref{prop:vds} reduces to
\begin{equation}\label{eq:pmf-spfti}
p(l) = C \min \{1,|l_\xi-\Nxi/2|^{-1} \cdot |\max\{l_{\rm x},l_{\rm y}\}|^{-1}\},
\end{equation}
where $C$ depends only on $\Nxi$ and $\Np$ and ensures that $\sum_{l}p(l)=1$.

Furthermore, under the requirements above, from Prop.~\ref{prop:vds} the estimation error achieved by $\Delta (\bs y^{\rm sp},p,\Omega,\bs \Phi_{\rm sp},\bs \Psi_{\rm sp})$ is bounded by
\begin{equation}\label{eq:estimation-error}\small
\|\bs x - \hat{\bs x}\| \le 2 K^{-1/2}\sigma_K(\bs \Psi^*_{\rm sp} \bs x) + \varepsilon.
\end{equation}
The following theorem summarizes this SP-FTI analysis.
\begin{thm}\label{thm:main-result}
Given $\epsilon\in (0,1]$ and $K \in \bb N$ such that $K \gtrsim \log(\Nhs)$ and 
\begin{equation}\small
M \gtrsim K \log(\Nxi)\log(\Np)\log(\epsilon^{-1}).
\end{equation}
Generate $M$ random (non-unique) indices associated with an index set $\Omega = \{\omega_1,\cdots,\omega_M\}$ such that $\omega_j \sim_\iid \beta$ for $j \in 
\range{M}$, with $\beta$ a \rv with the pmf in \eqref{eq:pmf-spfti}. Then, given the noisy compressive SP-FTI measurements $\bs y ^{\rm sp}$ in \eqref{eq:spfti cs acquisition}, the HS volume $\bs x$ can be approximated by solving $\Delta (\bs y^{\rm sp},p,\Omega,\bs \Phi_{\rm sp},\bs \Psi_{\rm sp})$ up to an error in \eqref{eq:estimation-error} with probability exceeding $1-\epsilon$.
\end{thm}
We observe that the optimum 1D pmf of selecting the rows of the matrix $\bs \Phi^*_{\rm sp}$ is linked to the 3D geometry of the sensing domain. In \eqref{eq:pmf-spfti}, the probability of programming the CA decreases inversely proportional to the magnitude of the OPD point, \ie its distance from the OPD origin. In addition, the probability of selecting a Hadamard pattern is inversely proportional to the maximum magnitude of spatial Hadamard ``frequencies''. For the sake of comparison, a uniform density sampling strategy, \ie $p(l) = \Nhs^{-1}$ for $l \in \range{\Nhs}$, and the sampling approach in \cite{jin2017hyperspectral} would require $M \gtrsim \Nhs K \log(\epsilon^{-1})$ and $M \gtrsim \Np K \log(\epsilon^{-1})$ measurements, respectively, which is equivalent to overexposing the observed object.
\section{Numerical results}                    
\label{sec:numerical-results}
We conduct several simulations to verify the performance of the proposed compressive SP-FTI on a simulated biological HS volume of size $(\Nxi,\Np) =(512, 64^2)$.  We construct it by mixing the coefficients of three spectral bands of a synthetic biological
RGB image selected from the benchmark images \cite{Ruusuvuori2008}, with the known spectra of three common fluorochromes \cite{fluorochromes}, see Fig.~\ref{fig:gt}~(right).
Compressive SP-FTI observations are formed according to \eqref{eq:spfti cs acquisition}, where the multiset $\Omega$ is randomly generated from the pmf \eqref{eq:pmf-spfti} and the variance of the associated Gaussian noise $\sigma^2_\Nyq$ is fixed w.r.t. to the desired Signal-to-Noise Ratio (SNR) reported in Fig.~\ref{fig:snr-curves}, with SNR $:= 10\log(\|\bs x\|^2/(\sigma_\Nyq^2 \Nhs ))$. This sensing procedure is repeated over 10 random realizations of the noise and the multiset $\Omega$. Fig.~\ref{fig:snr-curves} depicts the average Reconstruction SNR (RSNR) in dB with RSNR~$:=10\log(\|\bs x\|^2/\|\bs x - \hat{\bs x} \|^2)$. The value of $\varepsilon$ is computed via the empirical $95\jth$ percentile curve of the weighted noise power $\|\bs D \bs n\|$ over 100 Monte-Carlo realizations of the Gaussian noise and the index set $\Omega$. We recover HS volumes through $\Delta (\bs y^{\rm sp},p,\Omega,\bs \Phi_{\rm sp},\bs \Psi_{\rm sp})$ using SPGL1 toolbox~\cite{vandenBerg:2008}, referred to as CS reconstruction, and the Minimal Energy (ME) problem, \ie $\hat{\bs x}_{\rm me} := (\bs P_\Omega \bs \Phi^*_{\rm sp})^\dagger \bs y^{\rm sp}$, where $\dagger$ denotes the pseudo-inverse operator~\cite{candes2006robust}.

Poor performance of the ME solution (see dashed lines) highlights the necessity of leveraging HS sparsity prior. Since ME reconstruction does not consider the noise power, its performance does not change w.r.t. different SNR values. On the contrary, the solution of $\Delta (\bs y^{\rm sp},p,\Omega,\bs \Phi_{\rm sp},\bs \Psi_{\rm sp})$ is robust to the noisy measurements and is stable for compressible HS volumes. An example of the reconstructed HS volume w.r.t. $M/\Nxi  \approx (M+\Nxi)/(\Nhs + \Nxi) = 0.1$ is visualized in Fig.~\ref{fig:result-bands}. A comparison between these results and the ground truth values confirms that, even with 10\% of the SP-FTI measurements, the spatial and spectral content of the HS volume is successfully preserved.
\begin{figure}[t]
  \centering
  \begin{minipage}{0.49\linewidth}
    \centering
      \vspace{-4mm}
    \scalebox{0.4}{
%
%
\begin{tikzpicture}
\centering
\begin{axis}[%
width=7.5in,
height=2.5in,
at={(0in,0in)},
scale only axis,
point meta min=1,
point meta max=384,
axis on top,
xmin=0.5,
xmax=192.5,
y dir=reverse,
ymin=0.5,
ymax=64.5,
hide axis
]
\addplot [forget plot] graphics [xmin=0.5,xmax=192.5,ymin=0.5,ymax=64.5] {Fig_SpatialMap_GT-1.png};
\node[text=white, draw=none] at (rel axis cs:0.166,0.9) {\fontsize{22}{1}\selectfont $\bs{\nell_\nu = 72}$}; 
\node[text=white, draw=none] at (rel axis cs:0.5,0.9) {\fontsize{22}{1}\selectfont $\bs{\nell_\nu = 79}$}; 
\node[text=white, draw=none] at (rel axis cs:0.8333,0.9) {\fontsize{22}{1}\selectfont $\bs{\nell_\nu = 96}$}; 

 \draw [color=white,very thick] (64,0) -- (64,640);
 \draw [color=white,very thick] (129,0) -- (129,640);
\end{axis}
\end{tikzpicture}%
}
  \end{minipage}
  \begin{minipage}{0.49\columnwidth}
    \scalebox{0.4}{
%
%
\definecolor{mycolor1}{rgb}{0.00000,0.44700,0.74100}%
\definecolor{mycolor2}{rgb}{0.85000,0.32500,0.09800}%
\definecolor{mycolor3}{rgb}{0.92900,0.69400,0.12500}%
\begin{tikzpicture}
\centering
\begin{axis}[
width=6.8in,
height=2.5in,
at={(0in,0in)},
scale only axis,
xmin=1,
xmax=256,
xlabel={$\text{Wavenumber index (}\nell_\nu\text{)}$},
xtick={1,64,128,196,256},
xticklabels={1,64,128,196,256},
ticklabel style={font=\fontsize{22}{1}\selectfont},
ymin=-2,
ymax=102,
ylabel={$\text{Intensity}$},
ytick={0,100},
yticklabels={0,100},
xlabel style={at = {(0.5,-0.05)},font=\fontsize{30}{1}},ylabel style={at = {(0.01,0.5)},font=\fontsize{30}{1}},
axis background/.style={fill=white},
legend style={at={(1,0.98)},anchor=north east,legend cell align=left,align=left,fill=none,draw=none,font=\fontsize{25}{1}\selectfont}
]
\addplot [color=red,solid,line width=2.0pt]
  table[row sep=crcr]{%
1	0\\
2	0\\
3	0\\
4	0\\
5	0\\
6	0\\
7	0\\
8	0\\
9	0\\
10	0\\
11	0\\
12	0\\
13	0\\
14	0\\
15	0\\
16	0\\
17	0\\
18	0\\
19	0\\
20	0\\
21	0\\
22	0\\
23	0\\
24	0\\
25	0\\
26	0\\
27	0\\
28	0\\
29	0\\
30	0\\
31	0\\
32	0\\
33	0\\
34	0\\
35	0\\
36	0\\
37	0\\
38	0\\
39	0\\
40	0\\
41	0\\
42	0\\
43	0\\
44	0\\
45	0\\
46	0\\
47	0\\
48	0\\
49	0\\
50	0\\
51	0\\
52	0\\
53	0\\
54	0\\
55	0\\
56	0.425123840097398\\
57	1.01322132846114\\
58	1.43240469374638\\
59	1.92176290736576\\
60	2.49037897780377\\
61	3.06719923160167\\
62	4.05942633785257\\
63	5.76046766692309\\
64	9.0228210011953\\
65	13.6480579311096\\
66	17.3868053430139\\
67	19.8974910653507\\
68	22.0714950020302\\
69	27.7038877110484\\
70	38.3794873665128\\
71	62.3071575685072\\
72	100\\
73	99.7216816026445\\
74	47.2526528306809\\
75	16.5896717716924\\
76	5.70358052529219\\
77	2.19604585415567\\
78	0.8161982210223\\
79	0.328603355789523\\
80	0.131726016361397\\
81	0\\
82	0\\
83	0\\
84	0\\
85	0\\
86	0\\
87	0\\
88	0\\
89	0\\
90	0\\
91	0\\
92	0\\
93	0\\
94	0\\
95	0\\
96	0\\
97	0\\
98	0\\
99	0\\
100	0\\
101	0\\
102	0\\
103	0\\
104	0\\
105	0\\
106	0\\
107	0\\
108	0\\
109	0\\
110	0\\
111	0\\
112	0\\
113	0\\
114	0\\
115	0\\
116	0\\
117	0\\
118	0\\
119	0\\
120	0\\
121	0\\
122	0\\
123	0\\
124	0\\
125	0\\
126	0\\
127	0\\
128	0\\
129	0\\
130	0\\
131	0\\
132	0\\
133	0\\
134	0\\
135	0\\
136	0\\
137	0\\
138	0\\
139	0\\
140	0\\
141	0\\
142	0\\
143	0\\
144	0\\
145	0\\
146	0\\
147	0\\
148	0\\
149	0\\
150	0\\
151	0\\
152	0\\
153	0\\
154	0\\
155	0\\
156	0\\
157	0\\
158	0\\
159	0\\
160	0\\
161	0\\
162	0\\
163	0\\
164	0\\
165	0\\
166	0\\
167	0\\
168	0\\
169	0\\
170	0\\
171	0\\
172	0\\
173	0\\
174	0\\
175	0\\
176	0\\
177	0\\
178	0\\
179	0\\
180	0\\
181	0\\
182	0\\
183	0\\
184	0\\
185	0\\
186	0\\
187	0\\
188	0\\
189	0\\
190	0\\
191	0\\
192	0\\
193	0\\
194	0\\
195	0\\
196	0\\
197	0\\
198	0\\
199	0\\
200	0\\
201	0\\
202	0\\
203	0\\
204	0\\
205	0\\
206	0\\
207	0\\
208	0\\
209	0\\
210	0\\
211	0\\
212	0\\
213	0\\
214	0\\
215	0\\
216	0\\
217	0\\
218	0\\
219	0\\
220	0\\
221	0\\
222	0\\
223	0\\
224	0\\
225	0\\
226	0\\
227	0\\
228	0\\
229	0\\
230	0\\
231	0\\
232	0\\
233	0\\
234	0\\
235	0\\
236	0\\
237	0\\
238	0\\
239	0\\
240	0\\
241	0\\
242	0\\
243	0\\
244	0\\
245	0\\
246	0\\
247	0\\
248	0\\
249	0\\
250	0\\
251	0\\
252	0\\
253	0\\
254	0\\
255	0\\
256	0\\
};
\addlegendentry{\ R-PE (R-phycoerythrin)};

\addplot [color=green,solid,line width=2.0pt]
  table[row sep=crcr]{%
1	0\\
2	0\\
3	0\\
4	0\\
5	0\\
6	0\\
7	0\\
8	0\\
9	0\\
10	0\\
11	0\\
12	0\\
13	0\\
14	0\\
15	0\\
16	0\\
17	0\\
18	0\\
19	0\\
20	0\\
21	0\\
22	0\\
23	0\\
24	0\\
25	0\\
26	0\\
27	0\\
28	0\\
29	0\\
30	0\\
31	0\\
32	0\\
33	0\\
34	0\\
35	0\\
36	0\\
37	0\\
38	0\\
39	0\\
40	0\\
41	0\\
42	0\\
43	0\\
44	0\\
45	0\\
46	0\\
47	0\\
48	0\\
49	0\\
50	0\\
51	0\\
52	0\\
53	0\\
54	0\\
55	0\\
56	0\\
57	0\\
58	0\\
59	0\\
60	0\\
61	0\\
62	0\\
63	0\\
64	3.56436398700623\\
65	4.71793563336539\\
66	6.21968508989662\\
67	8.10283744740463\\
68	10.3797515327807\\
69	13.6129735250634\\
70	17.8965026542928\\
71	23.1972632121268\\
72	29.8445506581177\\
73	36.6165640659674\\
74	43.6186552430398\\
75	49.8499252645555\\
76	57.187936006264\\
77	67.3658510808578\\
78	80.3651424563954\\
79	93.3376595568273\\
80	100\\
81	92.3602658880212\\
82	68.9427660780757\\
83	41.0519785478685\\
84	18.5163272594056\\
85	7.16101544040108\\
86	2.36212019398192\\
87	0.793372866248487\\
88	0.431054487419222\\
89	0\\
90	0\\
91	0\\
92	0\\
93	0\\
94	0\\
95	0\\
96	0\\
97	0\\
98	0\\
99	0\\
100	0\\
101	0\\
102	0\\
103	0\\
104	0\\
105	0\\
106	0\\
107	0\\
108	0\\
109	0\\
110	0\\
111	0\\
112	0\\
113	0\\
114	0\\
115	0\\
116	0\\
117	0\\
118	0\\
119	0\\
120	0\\
121	0\\
122	0\\
123	0\\
124	0\\
125	0\\
126	0\\
127	0\\
128	0\\
129	0\\
130	0\\
131	0\\
132	0\\
133	0\\
134	0\\
135	0\\
136	0\\
137	0\\
138	0\\
139	0\\
140	0\\
141	0\\
142	0\\
143	0\\
144	0\\
145	0\\
146	0\\
147	0\\
148	0\\
149	0\\
150	0\\
151	0\\
152	0\\
153	0\\
154	0\\
155	0\\
156	0\\
157	0\\
158	0\\
159	0\\
160	0\\
161	0\\
162	0\\
163	0\\
164	0\\
165	0\\
166	0\\
167	0\\
168	0\\
169	0\\
170	0\\
171	0\\
172	0\\
173	0\\
174	0\\
175	0\\
176	0\\
177	0\\
178	0\\
179	0\\
180	0\\
181	0\\
182	0\\
183	0\\
184	0\\
185	0\\
186	0\\
187	0\\
188	0\\
189	0\\
190	0\\
191	0\\
192	0\\
193	0\\
194	0\\
195	0\\
196	0\\
197	0\\
198	0\\
199	0\\
200	0\\
201	0\\
202	0\\
203	0\\
204	0\\
205	0\\
206	0\\
207	0\\
208	0\\
209	0\\
210	0\\
211	0\\
212	0\\
213	0\\
214	0\\
215	0\\
216	0\\
217	0\\
218	0\\
219	0\\
220	0\\
221	0\\
222	0\\
223	0\\
224	0\\
225	0\\
226	0\\
227	0\\
228	0\\
229	0\\
230	0\\
231	0\\
232	0\\
233	0\\
234	0\\
235	0\\
236	0\\
237	0\\
238	0\\
239	0\\
240	0\\
241	0\\
242	0\\
243	0\\
244	0\\
245	0\\
246	0\\
247	0\\
248	0\\
249	0\\
250	0\\
251	0\\
252	0\\
253	0\\
254	0\\
255	0\\
256	0\\
};
\addlegendentry{\ Acridine orange};

\addplot [color=blue,solid,line width=2.0pt]
  table[row sep=crcr]{%
1	0\\
2	0\\
3	0\\
4	0\\
5	0\\
6	0\\
7	0\\
8	0\\
9	0\\
10	0\\
11	0\\
12	0\\
13	0\\
14	0\\
15	0\\
16	0\\
17	0\\
18	0\\
19	0\\
20	0\\
21	0\\
22	0\\
23	0\\
24	0\\
25	0\\
26	0\\
27	0\\
28	0\\
29	0\\
30	0\\
31	0\\
32	0\\
33	0\\
34	0\\
35	0\\
36	0\\
37	0\\
38	0\\
39	0\\
40	0\\
41	0\\
42	0\\
43	0\\
44	0\\
45	0\\
46	0\\
47	0\\
48	0\\
49	0\\
50	0\\
51	0\\
52	0\\
53	0\\
54	0\\
55	0\\
56	0\\
57	0\\
58	0\\
59	0\\
60	0\\
61	0\\
62	0\\
63	0\\
64	0\\
65	0\\
66	0\\
67	0\\
68	0\\
69	0\\
70	4.99148605175311\\
71	5.88630024495692\\
72	6.91390729429339\\
73	8.32812217454069\\
74	9.81896462230744\\
75	11.4466130001154\\
76	13.8979698304278\\
77	16.74794054043\\
78	19.5019547454361\\
79	22.1504297211369\\
80	25.3943703603723\\
81	28.7024739125831\\
82	31.5726640936457\\
83	35.0715848058799\\
84	38.9463545563847\\
85	43.0380526072911\\
86	46.773340110578\\
87	52.7067843754531\\
88	58.8285474053792\\
89	65.2377588631634\\
90	71.8967075062441\\
91	78.8404900055439\\
92	84.924867282287\\
93	90.2651954740089\\
94	94.6641522723048\\
95	97.5664475937293\\
96	99.6443820577769\\
97	100\\
98	99.0297467420538\\
99	96.1936527081008\\
100	92.26017238648\\
101	86.9530558811076\\
102	80.7340933529192\\
103	73.6319358819381\\
104	65.9289892779263\\
105	57.6817709129762\\
106	49.4431809172146\\
107	41.8696023749508\\
108	34.287474275664\\
109	26.5160119942333\\
110	19.2999390460755\\
111	13.3532295164691\\
112	8.47106244123438\\
113	5.19021408463166\\
114	3.17646389826742\\
115	2.00078332682424\\
116	1.37581883938494\\
117	1.08398814707413\\
118	0\\
119	0\\
120	0\\
121	0\\
122	0\\
123	0\\
124	0\\
125	0\\
126	0\\
127	0\\
128	0\\
129	0\\
130	0\\
131	0\\
132	0\\
133	0\\
134	0\\
135	0\\
136	0\\
137	0\\
138	0\\
139	0\\
140	0\\
141	0\\
142	0\\
143	0\\
144	0\\
145	0\\
146	0\\
147	0\\
148	0\\
149	0\\
150	0\\
151	0\\
152	0\\
153	0\\
154	0\\
155	0\\
156	0\\
157	0\\
158	0\\
159	0\\
160	0\\
161	0\\
162	0\\
163	0\\
164	0\\
165	0\\
166	0\\
167	0\\
168	0\\
169	0\\
170	0\\
171	0\\
172	0\\
173	0\\
174	0\\
175	0\\
176	0\\
177	0\\
178	0\\
179	0\\
180	0\\
181	0\\
182	0\\
183	0\\
184	0\\
185	0\\
186	0\\
187	0\\
188	0\\
189	0\\
190	0\\
191	0\\
192	0\\
193	0\\
194	0\\
195	0\\
196	0\\
197	0\\
198	0\\
199	0\\
200	0\\
201	0\\
202	0\\
203	0\\
204	0\\
205	0\\
206	0\\
207	0\\
208	0\\
209	0\\
210	0\\
211	0\\
212	0\\
213	0\\
214	0\\
215	0\\
216	0\\
217	0\\
218	0\\
219	0\\
220	0\\
221	0\\
222	0\\
223	0\\
224	0\\
225	0\\
226	0\\
227	0\\
228	0\\
229	0\\
230	0\\
231	0\\
232	0\\
233	0\\
234	0\\
235	0\\
236	0\\
237	0\\
238	0\\
239	0\\
240	0\\
241	0\\
242	0\\
243	0\\
244	0\\
245	0\\
246	0\\
247	0\\
248	0\\
249	0\\
250	0\\
251	0\\
252	0\\
253	0\\
254	0\\
255	0\\
256	0\\
};
\addlegendentry{\ TetraSpeck blue dye};

\end{axis}
\end{tikzpicture}%
}
  \end{minipage}
  \caption{Three spatial maps of ground truth HS volume (left); the known spectral signatures of three fluorochromes (right).}
  \label{fig:gt}
\end{figure}
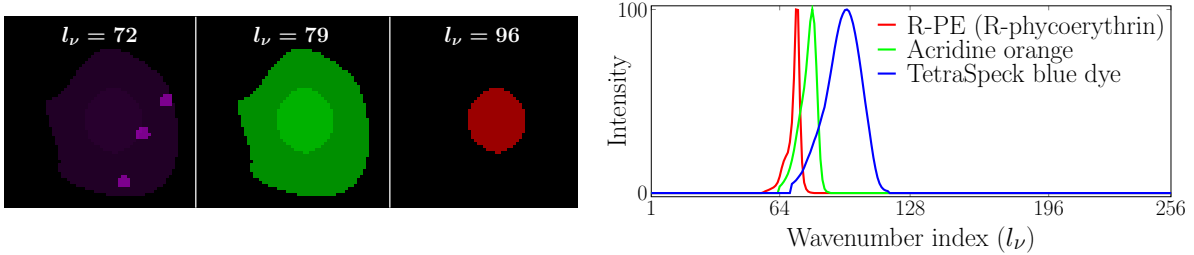
 \begin{figure}[t]
     \centering
     \scalebox{1}{\begin{tikzpicture}
\begin{axis}[%
width=3in,
height=2in,
at={(0.762in,0.486in)},
scale only axis,
xmin=0,
xmax=1,
xlabel={$\text{Measurement ratio}~(M/\Nhs)$},
xtick={0.1,0.2,0.3,0.4,0.5,0.6,0.7,0.8,0.9,1},
xticklabels={0.1,0.2,0.3,0.4,0.5,0.6,0.7,0.8,0.9,1},
xlabel style={at = {(0.5,0.02)},font=\fontsize{12}{1}\selectfont},
ticklabel style={font=\fontsize{10}{1}\selectfont},
ymin=0,
ymax=30,
ytick={0,5,10,15,20,25,30},
yticklabels={0,5,10,15,20,25,30},
ylabel={RSNR (dB)},
ylabel style={at = {(0.04,0.5)},font=\fontsize{12}{1}\selectfont},
axis background/.style={fill=white},
legend style={at={(0.95,0.05)},anchor=south east,legend cell align=left,align=left,fill=none,draw=none,font=\fontsize{6}{1}\selectfont},
legend entries={{SNR = 20 dB},{SNR = 15 dB},{SNR = 10 dB}}
]
\addlegendimage{no marks,black,solid,very thick},
\addlegendimage{no marks,red,solid,very thick},
\addlegendimage{no marks,mycolor1,solid,very thick},

\draw[red,line width=.5pt,anchor=center] (10,203) ellipse (4pt and 8pt);
\node[red,anchor=south]  at (10,215) {\fontsize{8}{1}\selectfont Fig.~\ref{fig:result-bands}(a)};

\draw[red,line width=.5pt,anchor=center] (10,100) ellipse (4pt and 8pt) ;
\node[red,anchor=north west]  at (11,100) {\fontsize{8}{1}\selectfont Fig.~\ref{fig:result-bands}(b)};

\draw[black,line width=.5pt,anchor=center] (80,165) ellipse (4pt and 4pt) ;
\node[black,anchor=north west]  at (80,163) {\fontsize{8}{1}\selectfont ME rec.};

\draw[black,line width=.5pt,anchor=center] (80,224) ellipse (4pt and 23pt) ;
\node[black,anchor=south west]  at (78,270) {\fontsize{8}{1}\selectfont CS rec.};

\node (e1) at (40,205) {};\node (e2) at (40,242) {};\draw[<->] (e1) -- (e2);
\node[align=left, text=black, draw=none,anchor=west]  at (40,224) {\fontsize{8}{1}\selectfont $\ \approx\text{4.5 dB}$};
\addplot [color=black,solid,line width=1.0pt,mark=o,mark options={solid}]
  table[row sep=crcr]{%
1	27.0009288372759\\
0.9	26.8044738031254\\
0.8	26.444183782717\\
0.7	26.0417174823581\\
0.6	25.5812991992637\\
0.5	25.0226708495979\\
0.4	24.4059693965304\\
0.3	23.5652143551228\\
0.2	22.4683043324788\\
0.1	20.3593133903976\\
0.02	9.6601896164235\\
};

\addplot [color=red,solid,line width=1.0pt,mark=+,mark options={solid}]
  table[row sep=crcr]{%
1	22.8604862094321\\
0.9	22.6419928287562\\
0.8	22.3784046688341\\
0.7	22.0127479639124\\
0.6	21.6156256211777\\
0.5	21.083391350909\\
0.4	20.4969344373692\\
0.3	19.7300658783017\\
0.2	18.6322825230232\\
0.1	16.9226131120466\\
0.02	8.9686703492764\\
};

\addplot [color=mycolor1,solid,line width=1.0pt,mark=triangle,mark options={solid,rotate=180}]
  table[row sep=crcr]{%
1	18.8036293628013\\
0.9	18.6142737278558\\
0.8	18.3025109074325\\
0.7	17.9215899656433\\
0.6	17.5056156280653\\
0.5	17.0176407110322\\
0.4	16.4235560562403\\
0.3	15.6878574381001\\
0.2	14.6844168436672\\
0.1	13.1887402325612\\
0.02	8.3133562468359\\
};

%

\addplot [color=black,dashed,line width=1.0pt]
  table[row sep=crcr]{%
1	17.1564465643498\\
0.9	16.9567677969346\\
0.8	16.7429348670392\\
0.7	16.5047658159974\\
0.6	16.1587349151421\\
0.5	15.6949215774402\\
0.4	15.1401098325\\
0.3	14.1775671531098\\
0.2	12.7517563645387\\
0.1	9.9964508826597\\
0.02	4.57715631696475\\
};

\addplot [color=mycolor1,dashed,line width=1.0pt,forget plot]
  table[row sep=crcr]{%
1	17.0662402976895\\
0.9	16.8638047410454\\
0.8	16.6451766562834\\
0.7	16.4007497853311\\
0.6	16.0521345033146\\
0.5	15.5859590744136\\
0.4	15.0272076851827\\
0.3	14.0727430568961\\
0.2	12.6606389389872\\
0.1	9.94798226889068\\
0.02	4.57280526382222\\
};
\addplot [color=red,dashed,line width=1.0pt,forget plot]
  table[row sep=crcr]{%
1	16.7840658462297\\
0.9	16.5731721683387\\
0.8	16.3395791147437\\
0.7	16.076404093863\\
0.6	15.7195146721981\\
0.5	15.2467977679993\\
0.4	14.6756774642417\\
0.3	13.746563521912\\
0.2	12.3787252227728\\
0.1	9.79761228630079\\
0.02	4.55907464403552\\
};

\end{axis}
\end{tikzpicture}%
}
     \caption{The performance of the proposed SP-FTI system.}
     \label{fig:snr-curves}
 \end{figure}
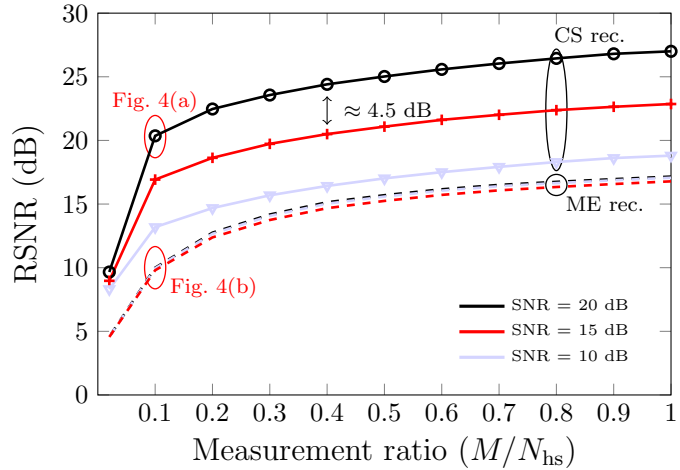
\begin{figure}[t!]
  \centering
  \begin{minipage}{0.49\linewidth}
    \centering
    \vspace{-1mm}
    \scalebox{0.4}{
%
%
\begin{tikzpicture}
\centering
\begin{axis}[%
width=7.5in,
height=2.5in,
at={(0in,0in)},
scale only axis,
point meta min=1,
point meta max=384,
axis on top,
xmin=0.5,
xmax=192.5,
y dir=reverse,
ymin=0.5,
ymax=64.5,
hide axis
]
\addplot [forget plot] graphics [xmin=0.5,xmax=192.5,ymin=0.5,ymax=64.5] {Fig_SpatialMap_CS_20dB-2.png};
\node[text=white, draw=none,anchor = north west] at (rel axis cs:0.01,0.95) {\fontsize{18}{1}\selectfont (a) CS rec.,~~~~~~~~~~~~~~~~~~~~~ RSNR = 20.21 dB}; 
\filldraw[color = white, white, very thick] (32,320)rectangle (34,340);
\filldraw[color = white, white, very thick] (96,320)rectangle (98,340);
\filldraw[color = white, white, very thick] (160,320)rectangle (162,340);
\node[above, align=center, text=white, draw=none] (b) at (32,320) {\fontsize{18}{1}\selectfont (32,32)};

 \draw [color=white,very thick] (64,0) -- (64,640);
 \draw [color=white,very thick] (129,0) -- (129,640);
\end{axis}
\end{tikzpicture}%
}
    \scalebox{0.4}{
%
%
\begin{tikzpicture}
\centering
\begin{axis}[%
width=7.5in,
height=2.5in,
at={(0in,0in)},
scale only axis,
point meta min=1,
point meta max=384,
axis on top,
xmin=0.5,
xmax=192.5,
y dir=reverse,
ymin=0.5,
ymax=64.5,
hide axis
]
\addplot [forget plot] graphics [xmin=0.5,xmax=192.5,ymin=0.5,ymax=64.5] {Fig_SpatialMap_ME_20dB-2.png};
\node[text=white, draw=none,anchor = north west] at (rel axis cs:0.01,0.95) {\fontsize{18}{1}\selectfont (b) ME rec.,~~~~~~~~~~~~~~~~~~~~~RSNR = 9.89 dB}; 
\filldraw[color = white, white, very thick] (32,320)rectangle (34,340);
\filldraw[color = white, white, very thick] (96,320)rectangle (98,340);
\filldraw[color = white, white, very thick] (160,320)rectangle (162,340);
 \draw [color=white,very thick] (64,0) -- (64,640);
 \draw [color=white,very thick] (129,0) -- (129,640);
\end{axis}
\end{tikzpicture}%
}
  \end{minipage}
  \begin{minipage}{0.49\linewidth}
    \centering
    \scalebox{0.4}{\input{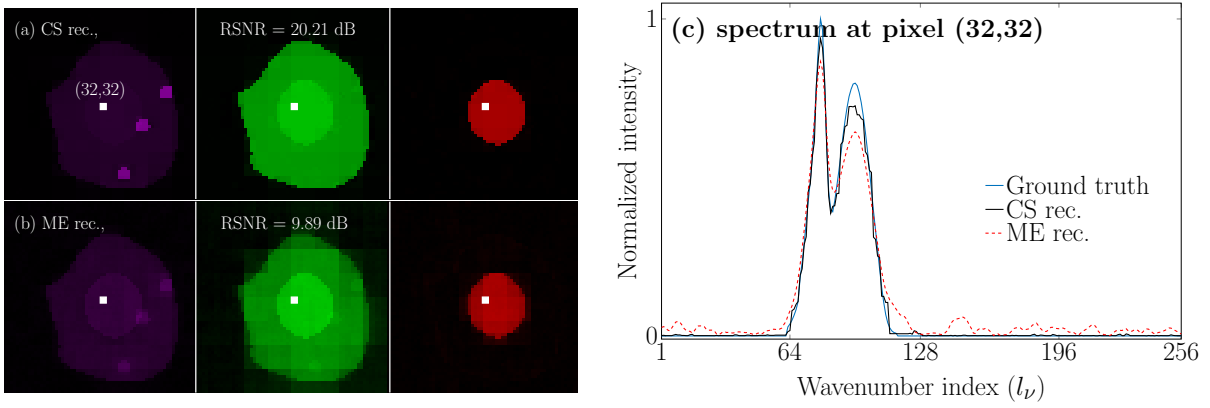}}
  \end{minipage}
  \caption{Three spatial maps of the reconstructed HS volumes (left); the spectral content at the centered pixel (right).}
  \label{fig:result-bands}
\end{figure}
\section{Conclusion}                             
\label{sec:conclusion}            
We have proposed a compressive SP-FTI where the light source can be optimally structured (using Hadamard patterns) in order to minimize the light exposure imposed on the observed object and the number of measurements. Our method is practically plausible without any hardware modification to the initial SP-FTI \cite{jin2017hyperspectral}. The theoretical recovery guarantee of this work supports any application of SP-FTI. However, extending this work to a biologically friendly SP-FTI, where the total amount of light exposure is assumed fixed will be the scope of a future work. Tracing the effect of other sparsity bases on our local coherence analysis, \eg the Daubechies wavelets~\cite{mallat2008wavelet}, is also postponed to future investigations.
\begin{center}
\vspace{1mm}
\textbf{APPENDIX: LOCAL COHERENCE BOUND}
\vspace{1mm}
\end{center}        
\label{sec:app}            

We here provide a proof sketch for computing the bound on the local coherence $\mu_l(\bs \Phi^*_{\rm sp}\bs \Psi_{\rm sp})$ for $l\in \range{\Nhs}$ in \eqref{eq:local-coherence-spfti}. We recall the relation $l = \Np (l_\xi -1)+ l_{\rm p}$, where $l_{\rm p} = \bar{N}_{\rm p}(l_{\rm y}-1)+l_{\rm x}$ between the 1D ``$l$''and the 3D ``$(l_\xi,l_{\rm x},l_{\rm y})$'' index representations.  Noting that $\bs \Phi^*_{\rm sp}\bs \Psi_{\rm sp}= (\bs \Phi_{\rm had}^*\bs \Psi_{\rm idhw}) \otimes (\bs \Phi_{\rm dft}^*\bs \Psi_{\rm dhw})$, the definition of local coherence \eqref{eq:local-coherence} gives
\begin{equation}\label{eq:app 1}
\mu_l(\bs \Phi^*\bs \Psi) = \mu_{l_\xi}(\bs \Phi_{\rm dft}^*\bs \Psi_{\rm dhw}) \cdot  \mu_{l_{p}}(\bs \Phi_{\rm had}^*\bs \Psi_{\rm idhw}).
\end{equation}
We have shown in \cite[Prop. 5.3]{moshtaghpour2018} that $\mu_{l_\xi}(\bs \Phi_{\rm dft}^*\bs \Psi_{\rm dhw}) \le \kappa^{\xi}_{l_\xi}:=\sqrt{2} \min\{1,|l_\xi-\Nxi/2|^{{-1/2}}\}$ and $\|\bs \kappa^{\xi}\|^2 \lesssim \log(\Nxi)$. We prove in this work (see below) that
$$
\mu_{l_{\rm p}}(\bs \Phi_{\rm had}^*\bs \Psi_{\rm idhw}) = \kappa^{\rm p}_{l_{\rm p}}:= \min\{1,2^{-\lfloor \log_2(\max\{l_{\rm x},l_{\rm y}\}-1) \rfloor}\},
$$
and $\|\bs \kappa^{\rm p}\|^2 = 1+3\log_2(\bar{N}_{\rm p})$. Thus, replacing these values in \eqref{eq:app 1} implies \eqref{eq:local-coherence-spfti}. Finally, we note $\|\bs \kappa\|^2 = \|\bs \kappa^{\xi}\|^2\cdot\|\bs \kappa^{\rm p}\|^2$.
\\\\
\textbf{Computation of} $\mu_{l_{\rm p}}(\bs \Phi_{\rm had}^*\bs \Psi_{\rm idhw})$: For fixed integers $r$, $\bar{N}_{\rm p} = 2^r$, and $\Np = \bar{N}^2_{\rm p}$, we first define 1-D dyadic wavelet levels as $\cl T_\ell:= \{l_{\ell-1}+1,\cdots,l_\ell\},~\ell \in \range{r}$ with $l_\ell := 2^\ell$ for $\ell \in \range{r}$ and $\cl T_0:=\{1\}$. In a matrix form, the 1D DHW basis in $\bb R^{ \bar{N}_{\rm p}\times \bar{N}_{\rm p}}$ can be constructed \cite{mallat2008wavelet,fino1972relations} as
		\begin{equation*}\small
		\bs \Psi_{\rm dhw} :=\bs W_{\bar{N}_{\rm p}} = \frac{1}{\sqrt{2}}\left[\bs W_{\bar{N}_{\rm p}/2} \otimes \begin{bmatrix} 1  \\ 1 \end{bmatrix},\bs I_{\bar{N}_{\rm p}/2} \otimes \begin{bmatrix} 1  \\ -1 \end{bmatrix}\right],
		\end{equation*}
with $\bs W_{1} = [1]$. Define also $\bs W^0_{1} := [1]$ and
		\begin{equation*}\small
		\bs \Psi^0_{\rm dhw} :=\bs W^0_{\bar{N}_{\rm p}} =\frac{1}{\sqrt{2}}\left[\bs W^0_{\bar{N}_{\rm p}/2} \otimes \begin{bmatrix} 1  \\ 1 \end{bmatrix},\bs I_{\bar{N}_{\rm p}/2} \otimes \begin{bmatrix} 1  \\ 1 \end{bmatrix}\right].
		\end{equation*}	
Up to a permutation of its columns, the 2D IDHW basis in $\bb R^{\Np \times \Np}$ \cite{mallat2008wavelet} can be constructed as $\bs \Psi_{\rm idhw} = [\bs \Psi_{0},\bs \Psi_{1}, \bs \Psi_{2}, \bs \Psi_{3}]$, where $\bs \Psi_0 := \bs 1_{\Np}$ and
		\begin{align*}
		\small
		& \bs \Psi_1 \!\!:= [\bs \Psi_{1,1},\cdots,\bs \Psi_{1,r}],~~ \bs \Psi_{1,\ell} \!\!:= (\bs \Psi^0_{\rm dhw} \bs P^\top_{\cl T_\ell})\! \otimes \!(\bs \Psi_{\rm dhw} \bs P^\top_{\cl T_\ell}),\\
		& \bs \Psi_2 \!\!:=  [\bs \Psi_{2,1},\cdots,\bs \Psi_{2,r}],~~ \bs \Psi_{2,\ell} \!\!:= (\bs \Psi_{\rm dhw} \bs P^\top_{\cl T_\ell})\! \otimes\! (\bs \Psi^0_{\rm dhw} \bs P^\top_{\cl T_\ell}),\\
		& \bs \Psi_3 \!\!:=  [\bs \Psi_{3,1},\cdots,\bs \Psi_{3,r}],~~ \bs \Psi_{3,\ell} \!\!:= (\bs \Psi_{\rm dhw} \bs P^\top_{\cl T_\ell})\! \otimes\! (\bs \Psi_{\rm dhw} \bs P^\top_{\cl T_\ell}),
		\end{align*}
with $\ell\! \in\! \range{r}$. The $\Np\! \times\! \Np$ Hadamard basis in Paley order~\cite{falkowski1996walsh,horadam2012hadamard} is defined by $\bs \Phi_{\rm had}\!:=\!\!\bs H_{\Np} \!\in\! \{\pm \Np^{-1/2}\}^{\Np \times\Np}$ where
		\begin{equation*}\small
		 \bs H_{\Np} := \frac{1}{\sqrt{2}}\left[\bs H_{\Np/2} \otimes \begin{bmatrix} 1  \\ 1 \end{bmatrix},\bs H_{\Np/2} \otimes \begin{bmatrix} 1  \\ -1 \end{bmatrix}\right]
		,\bs H_{1} := [1].
		\end{equation*}
Following these definitions, we can decompose the computation of local coherence as follows
	\begin{equation}\small \label{eq:app 2}
	\mu_{l_{\rm p}}(\bs H_\Np^* \bs \Psi_{\rm idhw}) = \max_{\substack{j = 1,2,3 \\ \ell=1,\cdots,r}} \mu_{l_{\rm p}}(\bs A_{j,\ell}) =:\kappa^{\rm p}_{l{\rm p}},
	\end{equation}
	where 
	\begin{align*}\small
	\bs A_{1,\ell} &:= (\bs H^*_{\bar{N}_{\rm p}} \bs W^0_{\bar{N}_{\rm p}} \bs P^\top_{\cl T_\ell}) \otimes (\bs H^*_{\bar{N}_{\rm p}} \bs W_{\bar{N}_{\rm p}} \bs P^\top_{\cl T_\ell}),\\
	\bs A_{2,\ell} &:= (\bs H^*_{\bar{N}_{\rm p}} \bs W_{\bar{N}_{\rm p}} \bs P^\top_{\cl T_\ell}) \otimes (\bs H^*_{\bar{N}_{\rm p}} \bs W^0_{\bar{N}_{\rm p}} \bs P^\top_{\cl T_\ell}),\\
	\bs A_{3,\ell} &:= (\bs H^*_{\bar{N}_{\rm p}} \bs W_{\bar{N}_{\rm p}} \bs P^\top_{\cl T_\ell}) \otimes (\bs H^*_{\bar{N}_{\rm p}} \bs W_{\bar{N}_{\rm p}} \bs P^\top_{\cl T_\ell}).
	\end{align*}
Beside, we obtain by direct computation
\begin{align*}\small
&\bs H^*_{\bar{N}_{\rm p}} \bs W_{\bar{N}_{\rm p}} = \begin{bmatrix} \bs H^*_{\bar{N}_{\rm p}/2} \bs W_{\bar{N}_{\rm p}/2} & \bs 0_{\bar{N}_{\rm p}/2 \times \bar{N}_{\rm p}/2} \\ \bs 0_{\bar{N}_{\rm p}/2 \times \bar{N}_{\rm p}/2} &  \bs H^*_{\bar{N}_{\rm p}/2} \end{bmatrix},\\
& \bs H^*_{\bar{N}_{\rm p}} \bs W^0_{\bar{N}_{\rm p}} = \begin{bmatrix} \bs H^*_{\bar{N}_{\rm p}/2} \bs W^0_{\bar{N}_{\rm p}/2} &  \bs H^*_{\bar{N}_{\rm p}/2} \\ \bs 0_{\bar{N}_{\rm p}/2 \times \bar{N}_{\rm p}/2} & \bs 0_{\bar{N}_{\rm p}/2 \times \bar{N}_{\rm p}/2} \end{bmatrix}.
\end{align*}
We can show by recursion that
\begin{align*}\small
&\bs P_{\cl T_t}\bs H_{2^r}^* \bs W_{2^r} \bs P^\top_{\cl T_\ell} = \begin{cases}
\bs H_{2^{t-1}}^*, & \text{if~} t=\ell\\
\bs 0_{|\cl T_t|\times |\cl T_\ell|}, & \text{otherwise}
\end{cases},\\
&\bs P_{\cl T_t}\bs H_{2^r}^* \bs W^0_{2^r} \bs P^\top_{\cl T_\ell} = \begin{cases}
\bs H_{2^{t-1}}^*, & \text{if~} t=\ell-1\\
\bs 0_{|\cl T_t|\times |\cl T_\ell|}, & \text{otherwise}
\end{cases}.
\end{align*}
This implies that $\mu_{j}(\bs H^*_{\bar{N}_{\rm p}} \bs W_{\bar{N}_{\rm p}} \bs P_{\cl T_\ell}^\top) = \min\{1,2^{-\lfloor \log_2(j-1)\rfloor/2}\}$, if $\lfloor \log_2(j-1)\rfloor = \ell-1$ (and zero otherwise), 
\begin{align*}\small
\mu_{j}(\bs H^*_{\bar{N}_{\rm p}} \bs W^0_{\bar{N}_{\rm p}} \bs P_{\cl T_\ell}^\top) = \begin{cases}
2^{-(\ell-1)/2}, & \text{if~} \lfloor \log_2(j-1)\rfloor \le \ell-2\\
 0, & \text{otherwise}
\end{cases}.
\end{align*}
We use again the relation $l_{\rm p} = \bar{N}_{\rm p}(l_{\rm y}-1)+l_{\rm x} \in \range{\Np}$, with $l_{\rm x}, l_{\rm y} \in \range{\bar{N}_{\rm p}}$, in order to decompose the computation of the local coherence of the matrices $\bs A_{j,\ell}$, as follows 
\begin{align*}\small
&\mu_{l_{\rm p}}(\bs A_{1,\ell}) = \mu_{l_{\rm x}}(\bs H^*_{\bar{N}_{\rm p}} \bs W_{\bar{N}_{\rm p}} \bs P^\top_{\cl T_\ell})\cdot  \mu_{l_{\rm y}}(\bs H^*_{\bar{N}_{\rm p}}\bs W^0_{\bar{N}_{\rm p}} \bs P^\top_{\cl T_\ell})\\
&\mu_{l_{\rm p}}(\bs A_{2,\ell}) = \mu_{l_{\rm x}}(\bs H^*_{\bar{N}_{\rm p}} \bs W^0_{\bar{N}_{\rm p}} \bs P^\top_{\cl T_\ell})\cdot  \mu_{l_{\rm y}}(\bs H^*_{\bar{N}_{\rm p}}\bs W_{\bar{N}_{\rm p}} \bs P^\top_{\cl T_\ell})\\
&\mu_{l_{\rm p}}(\bs A_{3,\ell}) = \mu_{l_{\rm x}}(\bs H^*_{\bar{N}_{\rm p}} \bs W_{\bar{N}_{\rm p}} \bs P^\top_{\cl T_\ell})\cdot  \mu_{l_{\rm y}}(\bs H^*_{\bar{N}_{\rm p}}\bs W_{\bar{N}_{\rm p}} \bs P^\top_{\cl T_\ell}).
\end{align*}
By individual computation of $\mu_{l_{\rm p}}(\bs A_{j,\ell})$ for $j =1,2,3$, \eqref{eq:app 2} gives $\mu_{l_{\rm p}}(\bs \Phi_{\rm had}^*\bs \Psi_{\rm idhw}) =\min\{1,2^{-\lfloor \log_2(\max\{l_{\rm x},l_{\rm y}\}-1) \rfloor}\}$ and $\| \bs \kappa^{\rm p}\|^2 = 1+ 3\log_2(\bar{N}_{\rm p})$, which completes the proof.
\vfil
\pagebreak
\bibliographystyle{IEEEtran}        
\bibliography{refs}
 
\end{document}